%% file: root.tex
\documentclass[letterpaper, 10 pt, conference]{ieeeconf}  % Comment this line out if you need a4paper
\usepackage{graphicx}
\newcommand{\methodname}{GazeAct}

\IEEEoverridecommandlockouts                              % This command is only needed if 
\usepackage{graphics} % for pdf, bitmapped graphics files
\usepackage{epsfig} % for postscript graphics files
\usepackage{mathptmx} % assumes new font selection scheme installed
\usepackage{times} % assumes new font selection scheme installed
\usepackage{amsmath} % assumes amsmath package installed
\usepackage{amssymb}  % assumes amsmath package installed
\usepackage{booktabs} % for \toprule, \midrule, \bottomrule
\usepackage{pifont}
\usepackage{multirow} % for \multirow in tables
\usepackage[table]{xcolor}
\definecolor{gazerowgray}{gray}{0.94}
\definecolor{checkgreen}{RGB}{0,130,0}
\definecolor{crossred}{RGB}{180,0,0}

\newcommand{\cmark}{\textcolor{checkgreen}{\ding{51}}}
\newcommand{\xmark}{\textcolor{crossred}{\ding{55}}}

\usepackage{xspace}
\usepackage{tabularx}
\usepackage{rotating}

\usepackage[table]{xcolor}
\definecolor{avggray}{gray}{0.94}

\usepackage[table]{xcolor}
\definecolor{avgrowgray}{gray}{0.94}

\title{\LARGE \bf
GazeAct: A Human-Gaze-Annotated Manipulation Dataset for Attention-Regularized Policies
}

 \author{Anupam Pani$^{1}$, Wang Yuan$^{1}$, Qifan Wang$^{1}$, Yanchao Yang$^{1,2}$% <-this % stops a space
\thanks{$^{1}$ HKU Musketeers Foundation Institute of Data Science}%
\thanks{$^{2}$Department of Electrical and Electronic Engineering, The University of Hong Kong}%
}

\begin{document}

\maketitle
\thispagestyle{empty}
\pagestyle{empty}

\input{sections/abstract}
\input{sections/introduction}
\input{sections/related_work}

\input{sections/method}

\input{sections/experiments}

\input{sections/conclusion}
\bibliographystyle{IEEEtran}
\bibliography{IEEEfull,references}

\end{document}

%% file: sections/abstract.tex
\begin{abstract}

Vision-language-action (VLA) policies must ground language instructions in
the right objects, contact points, and target locations for reliable manipulation.
Human gaze provides a natural signal for this grounding, but existing robot
datasets rarely pair real human gaze with the multi-camera observations used by
modern policies. We introduce \methodname{}, a gaze-equipped real-robot
manipulation dataset and training framework for multi-view VLA policies.
Using screen-based teleoperation, we record gaze directly on the robot camera
feed viewed by the operator, yielding $12{,}000$ demonstrations across $40$
tasks and five domains with $300\,\mathrm{Hz}$ gaze synchronized with images,
proprioception, and actions. We transfer gaze from the operator-selected view
to the remaining synchronized cameras, producing spatially consistent targets
across the policy's visual inputs. We then use these gaze-derived targets to
regularize language-conditioned visual attention during training while keeping
inference gaze-free. Across $18$ real-world tasks, our method improves average
success from $48.1\%$ to $58.6\%$, with gains up to $25$ percentage points.
Ablations show that gaze works best as soft supervision and that real human
gaze provides stronger attention targets than generic attention-shaping priors.
The gains transfer to LIBERO-Spatial ($57.9\%$ to $69.6\%$) and different VLA
backbones.These results show that gaze collected in the robot observation space can provide
effective training-time targets for improving visual grounding in VLA
manipulation.

\end{abstract}

%% file: sections/introduction.tex
\section{Introduction}
\label{sec:intro}

VLA models have become a promising paradigm for
language-conditioned robotic manipulation, mapping visual observations, language
instructions, and proprioception to continuous actions~\cite{
brohan2023rt2visionlanguageactionmodelstransfer,
intelligence2025pi05visionlanguageactionmodelopenworld,
kim2024openvlaopensourcevisionlanguageactionmodel}. Despite large-scale
pretraining, a basic difficulty remains: knowing \emph{where} to look. Many
manipulation failures arise not from misunderstanding the language instruction
alone, but from failing to attend to small, transient, and ambiguous visual
regions such as the object, grasp region, contact point, handle, button, or
target location. Standard VLA policies must infer these regions indirectly from
action supervision.

Human gaze provides a natural signal for this missing visual grounding. During
manipulation, demonstrators often look at task-relevant objects 
and target regions before or while acting. Gaze can therefore provide a compact
trace of the visual evidence used to support action decisions. However, the
primary obstacle is not the idea of using gaze, but the lack of robot learning
data in which gaze is aligned with the observations consumed by modern policies.
Large robot manipulation datasets typically contain images, proprioception, and
actions, but no human gaze. Conversely, existing gaze-equipped manipulation
datasets usually record gaze in a single egocentric, head-mounted, or displayed
view, which does not directly provide supervision over the multi-view robot
observations used by current VLA policies.

We introduce \methodname{}, a gaze-annotated manipulation dataset and acquisition
pipeline designed to close this alignment gap. Our key idea is to record gaze
directly on the robot camera stream used by the teleoperator, so each gaze point
is obtained in robot-image coordinates and tied to the visual evidence used for
action selection. We implement this with a screen-based ALOHA teleoperation
system in which a Tobii Pro Spectrum eye tracker records gaze on the active
robot camera view while the robot logs multi-view images, proprioception, and
actions. Hardware triggers and synchronized clocks align gaze and robot streams
in time, while active-view logging records which camera the operator observed at
each moment. Because only one view is displayed at a time but VLA policies
consume multi-view observations, we transfer gaze to the remaining synchronized
camera views, producing spatially consistent gaze targets for all visual inputs
used by the policy.

\begin{figure*}[t]
\vspace{1mm}
\centering
\includegraphics[width=0.96\linewidth]{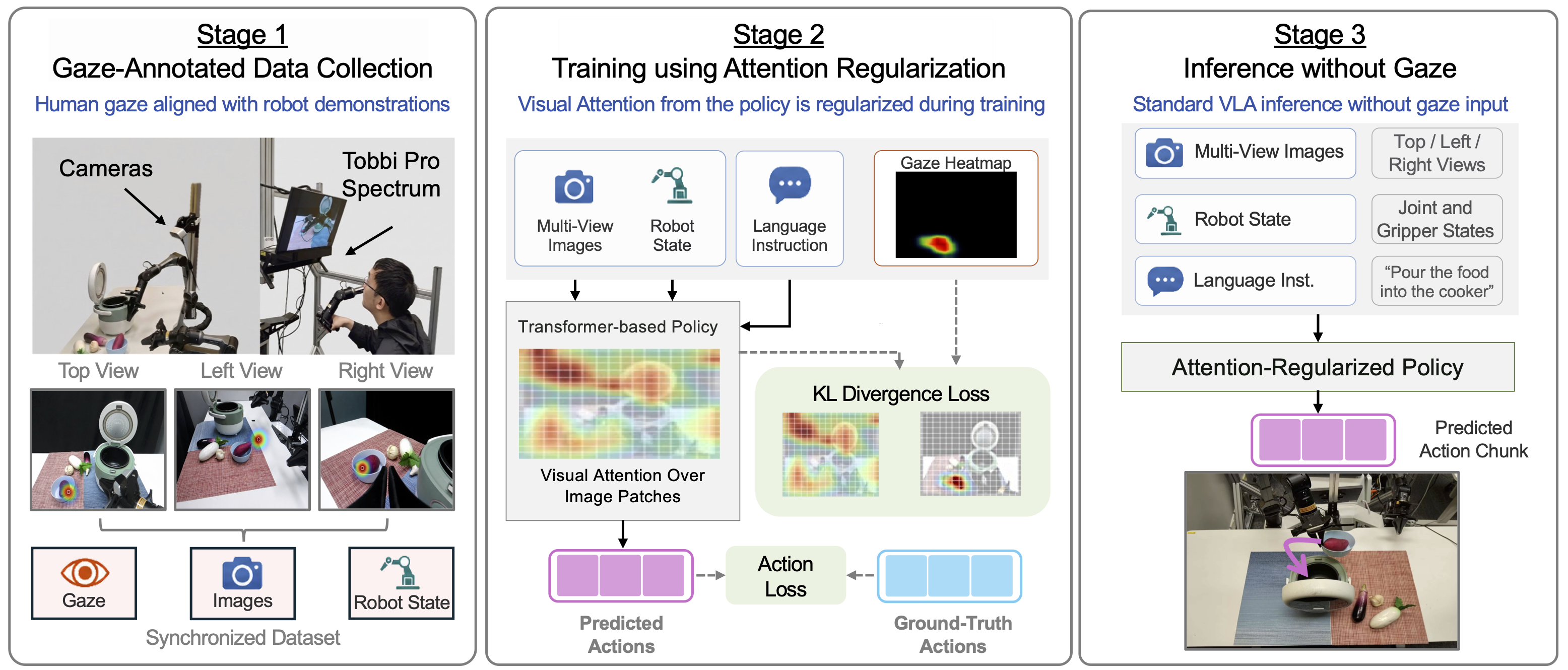}
\vspace{-4mm}
\caption{\textbf{Overview of \methodname{}.}
\emph{Left:}
We collect teleoperation demonstrations with synchronized robot observations,
proprioception, actions, and eye gaze recorded on the active robot camera view.
\emph{Middle:}
Recorded gaze is first aligned to the remaining camera views and converted into
patch-level gaze distributions. During training, these distributions regularize
the policy's language-conditioned visual attention toward task-relevant regions.
\emph{Right:}
At inference, gaze is not required as a signal and the policy predicts actions from standard VLA
inputs: images, language, and proprioception.
}
\label{fig:teaser}
\vspace{-6pt}
\end{figure*}

The resulting dataset contains $12{,}000$ demonstrations across $40$
manipulation tasks in five domains, with synchronized gaze, multi-view RGB
observations, proprioception, and actions. To turn this gaze into a learning
signal, we convert synchronized and cross-view-aligned gaze into patch-level
distributions over the policy's visual tokens. We then use these gaze-derived distributions as targets for attention
regularization, encouraging the policy's language-conditioned visual attention
to align with human-fixated regions during training., without changing the policy architecture or
inference-time inputs.

Real human gaze therefore acts as an auxiliary supervision signal: it guides visual grounding during
training, but is removed at deployment. We evaluate this supervision signal through
expanded real-world rollouts, robustness trials, gaze-source and
regularization-strength ablations, transfer across VLA backbones, and simulation
transfer. Across these settings, real robot-view gaze improves policy learning
most strongly when manipulation depends on grounding small objects, contact
points, or target regions.

Our contributions are threefold. \emph{First}, we introduce \methodname{}, a
gaze-annotated manipulation dataset in which human gaze is grounded in the
robot's own observations and made available across synchronized camera views.
\emph{Second}, we develop a screen-based, hardware-synchronized data-collection
and cross-view alignment pipeline for pairing high-frequency human gaze with
multi-view robot demonstrations. \emph{Third}, we show that gaze-derived targets provide an effective training-time
signal for regularizing visual attention in modern VLA policies,
improving visual grounding and manipulation success while preserving the
standard gaze-free deployment interface.

%% file: sections/related_work.tex
\section{Related Work}
\label{sec:related}

\textbf{Manipulation Datasets with and without Gaze.}
Large-scale teleoperation datasets have driven the recent generation of VLA models. 
RT-1~\cite{brohan2023rt1roboticstransformerrealworld},
BridgeData~V2~\cite{walke2024bridgedatav2datasetrobot} and DROID~\cite{khazatsky2025droidlargescaleinthewildrobot}
provide hundreds of thousands of demonstrations across diverse robots and scenes, but record only what the operator did, 
not where they looked. 
A smaller body of work adds human gaze. 
HARMONIC~\cite{newman2020harmonicmultimodaldatasetassistive} captures gaze
with head-mounted glasses in a single assistive-eating task, but projects it onto the operator's egocentric view rather than the robot's cameras.
DAA~\cite{kim2024multitaskrealrobotdatagaze} and
GazeBot~\cite{takizawa2025enhancingreusabilitylearnedskills} record real gaze on the robot's stereo feed through an HMD (Head-mounted display) at large scale, but the HMD shows a single stereo composite, so gaze covers only one view and trajectories last a few seconds, 
capturing brief reaches rather than complete manipulations. In contrast, \methodname{} records
screen-based gaze directly on the robot camera feed used for teleoperation,
synchronizes it with robot observations and actions, and transfers it across
camera views so that gaze can supervise the multi-view inputs used by VLA
policies.

\textbf{Gaze as a Learning Signal.}
Human gaze has been used in robot learning and visuomotor learning in several
ways. One line of work uses gaze as an offline cue for segmenting demonstrations;
for example, Takizawa et
al.~\cite{takizawa2025gazeguidedtaskdecompositionimitation} use fixation
transitions to decompose long-horizon demonstrations into sub-tasks. A second
line modifies the perceptual input pipeline, using gaze or predicted gaze to crop,
zoom, or foveate observations~\cite{Kim_2021,li2024virt,chuang2025lookfocusactefficient}.
These approaches can improve perception, but often change the input interface or
require gaze-like predictions at deployment. A third line, closer to our work,
uses gaze only during training to guide internal attention. Prior work has shown
that aligning model saliency or attention with human gaze can improve imitation
learning and visual recognition while keeping inference gaze-free~\cite{
saran2021efficientlyguidingimitationlearning,
thammineni2020selectiveeyegazeaugmentationenhance,
zhang2018agillearningattentionhuman,
zhang2019atariheadatarihumaneyetracking,
li2025gazeguidedlearningavoidingshortcut,
rong2021humanattentionfinegrainedclassification,
alacam-etal-2024-eyes,
chen2025gazeinsightbridginghuman,
khaertdinova2024gazeassistedmedicalimagesegmentation}.
In vision-language models, Voila-A~\cite{yan2023voilaaaligningvisionlanguagemodels}
uses gaze heatmaps to modulate attention, while gaze-regularized
models~\cite{panigaze} align attention with ground-truth gaze during training.
\methodname{} brings this training-time gaze supervision paradigm to real
teleoperated VLA manipulation, using synchronized robot-view gaze, cross-view
alignment, and gaze-free deployment.

% %%\vspace{-2mm}
\textbf{Vision-Language-Action Models and Attention.}
Our method builds on transformer-based VLA policies, including
$\pi_{0.5}$~\cite{intelligence2025pi05visionlanguageactionmodelopenworld},
RT-1~\cite{brohan2023rt1roboticstransformerrealworld},
OpenVLA~\cite{kim2024openvlaopensourcevisionlanguageactionmodel},
RoboMamba~\cite{liu2024robomambaefficientvisionlanguageactionmodel},
Octo~\cite{octomodelteam2024octoopensourcegeneralistrobot}, and
SAM2Act~\cite{fang2025sam2actintegratingvisualfoundation}. 
These models map
visual observations and language instructions to robot actions through internal
vision-language representations, often including attention over visual tokens.
\methodname{} uses this attention interface as the point of supervision: real
human gaze is converted into patch-level gaze distributions and used to
regularize language-conditioned visual attention during training. Because the
regularizer acts on the policy's internal attention rather than changing the
input stream or action head, it can be applied to different VLA backbones while
preserving the standard inference interface.

%% file: sections/method.tex
\section{Method}
\label{sec:method}
\methodname{} provides a framework for converting operator gaze, recorded on the
robot camera feed during teleoperation, into training-time supervision
for multi-view VLA policy learning. The
framework has three components. First, we collect a gaze-equipped teleoperation
dataset in which human gaze is recorded directly on the robot camera and synchronized with images, proprioception, and actions
(Section~\ref{sec:dataset}). Second, because the operator views one camera stream
at a time while the policy consumes multi-view observations, we transfer each
recorded gaze point from the active view to the remaining synchronized robot
views. This extends gaze supervision beyond a single displayed camera and
produces patch-level gaze distributions over all the visual inputs used by the
policy.
Third, we use attention regularization 
(Section~\ref{sec:reg}) to convert these distributions into an auxiliary
training signal that encourages language-conditioned visual attention to align
with human-fixated, task-relevant regions. Gaze is used only to construct training-time supervision targets; at inference,
the policy receives the same images, language, and proprioception as the
baseline VLA policy.

\begin{figure}[t]
  \centering
  \includegraphics[width=0.85\linewidth]{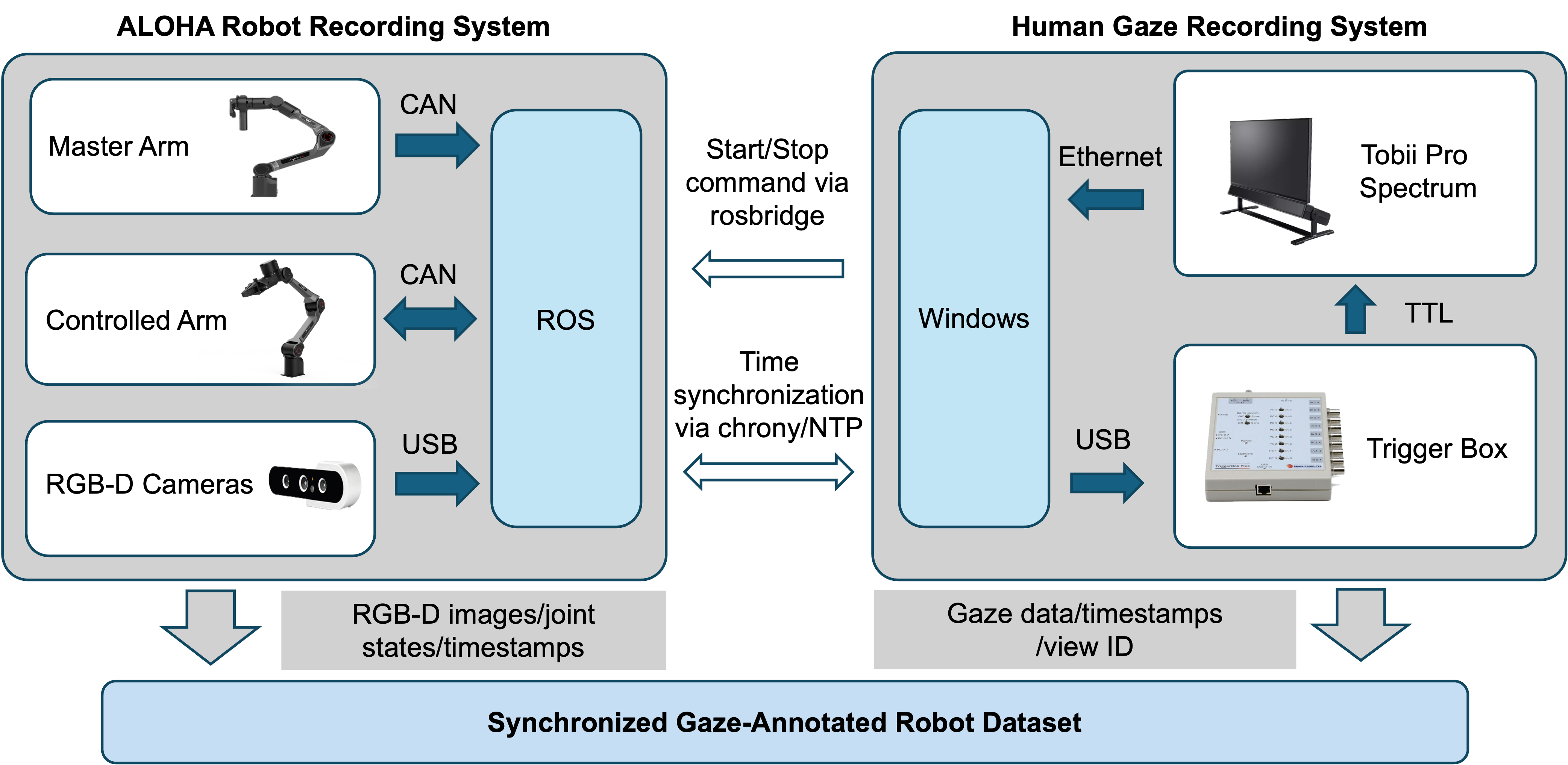}
  \vspace{-3mm}
\caption{
\textbf{Data collection and synchronization system.}
During ALOHA teleoperation, robot observations, proprioception, and actions are
logged on the robot side, while a screen-based eye tracker records the
operator's gaze on the displayed camera feed. Hardware triggers define
shared trial boundaries, and synchronized timestamps support post hoc
alignment of gaze with images, proprioception, and actions.}
  \label{fig:data_collection_equipment}
  \vspace{-3mm}
\end{figure}

% =====================================================================
%  3.1 PROBLEM FORMULATION
% =====================================================================
\subsection{Problem Formulation}
\label{sec:problem}

We formulate manipulation as conditional action prediction with a VLA policy. At
timestep $t$, the policy observes multi-view RGB images, a language instruction,
and proprioception,
\begin{equation}
o_t = \{\, I_t^{1:n},\; \ell,\; q_t \,\},
\end{equation}
where $I_t^{1:n}=\{I_t^1,\dots,I_t^n\}$ are images from $n$ camera views,
$\ell=[w_1,\dots,w_T]$ is the tokenized instruction, and
$q_t\in\mathbb{R}^{d_p}$ is proprioceptive state such as joint positions and
gripper state. The policy predicts a short-horizon action chunk,
\begin{equation}
A_t = [\,a_t, a_{t+1}, \dots, a_{t+h-1}\,], \qquad h=50,
\end{equation}
and is trained to model the conditional distribution of future actions,
$p_\theta(A_t \mid o_t)$, where $\theta$ denotes the policy parameters optimized
from teleoperated image-action demonstrations.
In addition to these standard imitation-learning signals, \methodname{} provides
synchronized human gaze during training. We represent this gaze as patch-level
distributions $\{G_t^i\}_{i=1}^{n}$ over the policy's visual tokens and use them for regularizing language-conditioned visual attention.

% --------------------------- FIGURE (CROSS-VIEW) --------------------
\begin{figure}[t]
  \centering
  \includegraphics[width=0.98\linewidth]{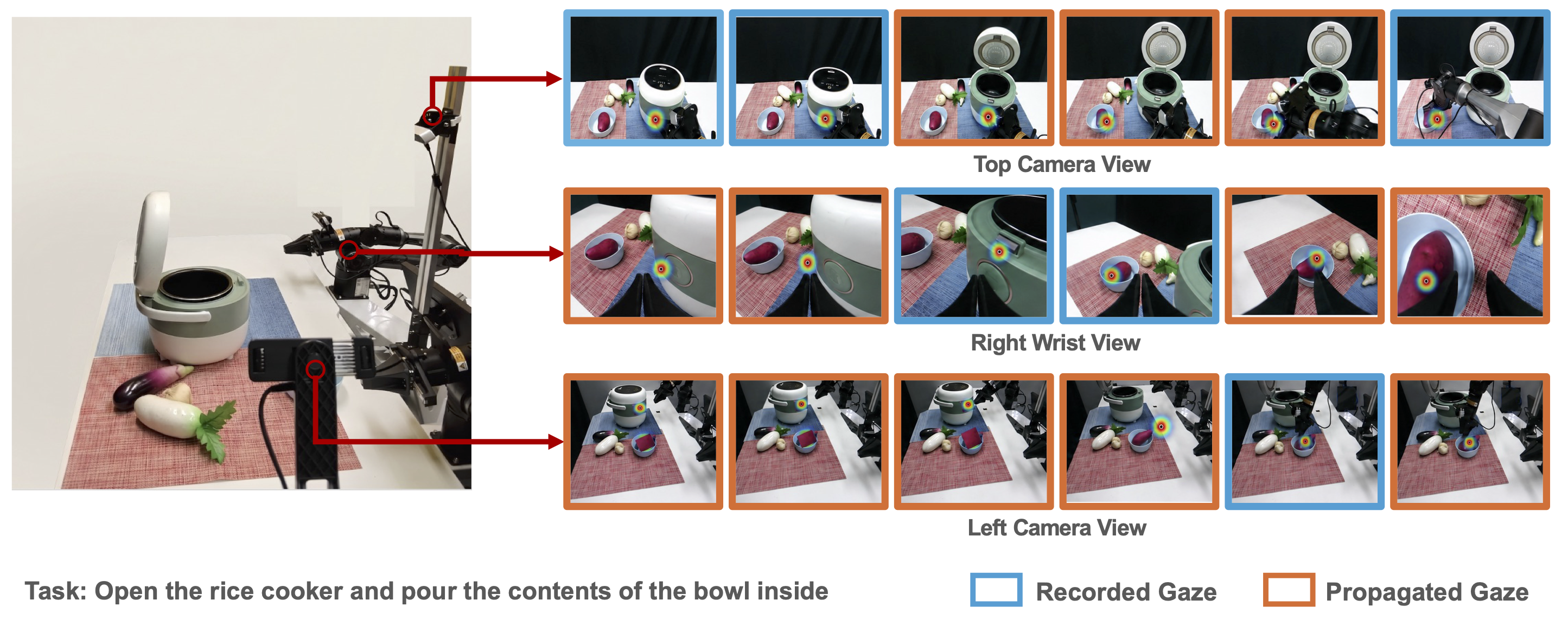}
  \vspace{-3mm}
  \caption{
\textbf{Cross-View Gaze Alignment for Multi-View Robot Observations.}
We visualize a teleoperated trajectory for the task ``open the rice cooker and
pour the contents of the bowl inside.'' At each timestep, gaze is recorded only
on the active camera view (blue edges); our cross-view alignment propagates it
to the remaining synchronized views (orange edges). The overlays show that the
propagated gaze remains spatially consistent across viewpoints and
task-relevant regions.}
  \label{fig:crossview}
    \vspace{-12pt}
\end{figure}

% --------------------------- FIGURE (ARCH) --------------------------
\begin{figure*}[t]
  \centering
  \includegraphics[width=0.98\linewidth]{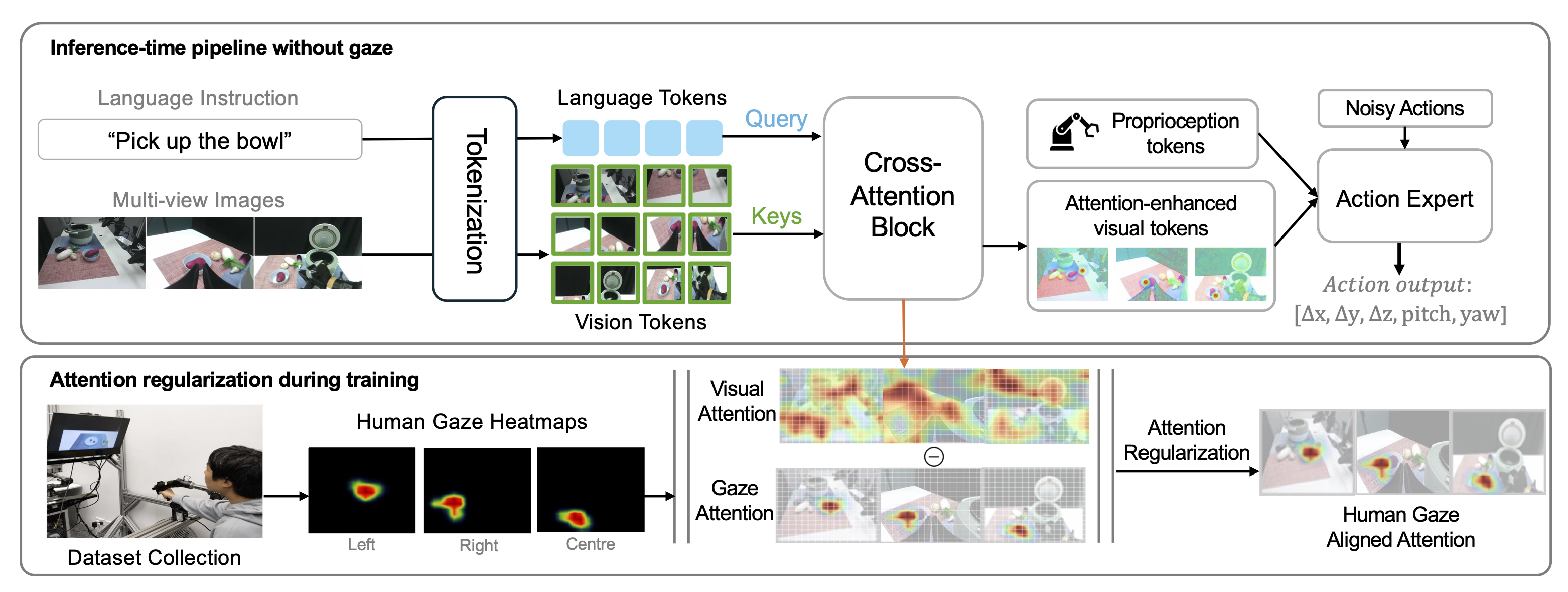}
  \vspace{-4mm}
\caption{
\textbf{Attention Regularization.}
\emph{Top:} the transformer-based policy processes language and image patch
tokens, producing visual attention over robot observations while learning to
predict actions.
\emph{Bottom:} during training, recorded teleoperation gaze is converted into
patch-level heatmaps and used as an auxiliary signal to guide the
policy's visual attention toward human-fixated, task-relevant regions. At
inference, gaze is not required and the policy predicts actions from standard
VLA inputs.
}
\vspace{-4mm}
  \label{fig:arch}
\end{figure*}

\subsection{Gaze-Equipped Teleoperation Dataset}
\label{sec:dataset}

\textbf{Data Collection.}
We collect demonstrations on an ALOHA teleoperation platform with a screen-based
visual feedback interface (Figure~\ref{fig:data_collection_equipment}). During
each demonstration, the operator controls the robot while observing the active
camera stream on a Tobii Pro Spectrum eye-tracking display, which records gaze at
$300\,\mathrm{Hz}$. Because gaze is recorded directly on the displayed robot
camera feed, each gaze point is expressed in robot-image coordinates and reflects
the visual evidence used for action selection. The operator can switch among
three views -- left wrist, right wrist, and a fixed top view -- to support
different manipulation stages, from scene understanding to grasp alignment and
placement. Each gaze sample is logged with a timestamp, an image-normalized 2D
coordinate, and the active camera view. The dataset spans $40$ manipulation tasks across five domains: \emph{home},
\emph{hospital}, \emph{kitchen}, \emph{laboratory}, and \emph{sports}. We collect
demonstrations from $20$ participants, each contributing $15$ demonstrations per
task, for $300$ demonstrations per task and $12{,}000$ demonstrations in total.

\textbf{Synchronization.}
% Gaze, camera, proprioception, and action streams are recorded by separate
% devices and at different rates, so \methodname{} uses explicit synchronization
% to align them within each demonstration. Hardware TTL triggers define shared
% trial boundaries: a start trigger begins both gaze recording and robot-side
% logging, and a matching stop trigger closes the same trial across systems.
% Within these boundaries, chrony/NTP-synchronized clocks support post hoc
% alignment of each $300\,\mathrm{Hz}$ gaze sample to the nearest robot timestep
% and camera frame. We retain the full gaze stream, allowing brief fixations and
% rapid gaze shifts around grasping, contact, and view switching to contribute to
% the gaze heatmaps used for supervision. Figure~\ref{fig:data_collection_equipment}
% shows the data collection equipment and synchronized recording setup.
Gaze, camera, proprioception, and action streams are recorded by separate
devices and at different rates, so \methodname{} uses explicit synchronization
to align them within each demonstration. Hardware TTL (Transistor-Transistor Logic) triggers define shared
trial boundaries: a start trigger begins both gaze recording and robot-side
logging, and a matching stop trigger closes the same trial across systems.
Within these boundaries, chrony/NTP-synchronized clocks support post hoc
alignment of each $300\,\mathrm{Hz}$ gaze sample to the nearest robot timestep
and camera frame. We retain the full gaze stream, allowing brief fixations and
rapid gaze shifts around grasping, contact, and view switching to contribute to
the gaze heatmaps used for supervision. Figure~\ref{fig:data_collection_equipment}
shows the data collection equipment and synchronized recording setup.

\textbf{Cross-View Gaze Alignment.}
Gaze is recorded on the camera view displayed to the operator, but the VLA policy
receives multi-view observations. To use gaze as supervision for all visual
inputs, \methodname{} transfers each active-view gaze point to the remaining
synchronized camera views. This extends real human gaze beyond a single displayed
view and makes it usable as multi-view privileged supervision.

% Before data collection, we calibrate the camera system using a chessboard target
% to obtain camera intrinsics and extrinsics. Given a gaze point in the active
% source view, we first compute a geometry-based candidate in each target view by
% back-projecting the source gaze point using depth, transforming the resulting 3D
% point through the calibrated robot frame, and projecting it into the target
% camera. This produces a physically grounded estimate of where the source-view
% gaze should appear in the target image.

Before data collection, we calibrate the multi-camera system using a chessboard
target to estimate each camera's intrinsic parameters and its pose relative to
the robot frame. This calibration places the fixed top camera and wrist cameras
in a shared geometric coordinate system. Given a gaze point in the active source
view, we first compute a geometry-based candidate in each target view by
back-projecting the source gaze point using depth, transforming the resulting
3D point through the calibrated robot frame, and projecting it into the target
camera. The resulting candidate provides a physically grounded estimate of where
the source-view gaze should appear in each synchronized target image.

Because geometry-only transfer can shift under depth noise, calibration error,
or wrist-camera pose error, we further refine the transferred point using local
image correspondences. Specifically, the geometry-based candidate defines a
bounded search region in the target view, and LightGlue~\cite{lindenberger2023lightglue}
is used to match local visual features between the source and target images.
Reliable matches inside this geometry-guided region refine the transferred gaze
point. This combines the physical consistency of calibrated geometry with the
local correction ability of feature matching.

When cross-view transfer is unreliable, for example because depth is missing,
the projected point falls outside the target image, or the views have
insufficient overlap, the method falls back to a center prior. Because
\methodname{} retains the raw source-view gaze together with synchronized
multi-view camera streams, the transfer step can be recomputed with improved
registration methods without recollecting demonstrations. An example of
cross-view gaze transfer is shown in Figure~\ref{fig:crossview}.

\textbf{Heatmap and Patch-Level Distribution.}
Each valid gaze coordinate is converted into a spatial heatmap $H_t^{i}$ on view
$i$ by placing a Gaussian kernel at the gaze location, aggregated over a short
temporal window to suppress micro-saccades and frame-level jitter. Since the policy attends over visual tokens rather
than pixels, we pool each heatmap into a \emph{patch-level distribution} matching
the visual-token grid. Let $\mathcal{P}=\{\mathbf{P}_j\}_{j=1}^{N_v}$ denote the
patch partition corresponding to the $N_v$ visual tokens. The gaze mass on patch
$\mathbf{P}_j$ is
\begin{equation}
G_{t,j}^{i} = \frac{1}{Z}\sum_{(x,y)\in\mathbf{P}_j} H_t^{i}(x,y),
\qquad Z = \sum_{x,y} H_t^{i}(x,y).
\label{eq:pool}
\end{equation}
The resulting vector $G_t^{i}=[G_{t,1}^{i},\dots,G_{t,N_v}^{i}]$ is normalized so
that $\sum_{j=1}^{N_v}G_{t,j}^{i}=1$, giving a gaze-derived target distribution
over the visual tokens of view $i$ for the attention regularization objective in
Section~\ref{sec:reg}.

\begin{table*}[t]
\caption{
Comparison of \methodname{} with representative robot manipulation and
gaze-augmented datasets. \textbf{Recorded Human Gaze} denotes whether real gaze
is collected; \textbf{Robot-View Gaze} whether gaze is recorded on robot camera
observations; \textbf{Multi-View Gaze} whether gaze is available across multiple
camera views; and \textbf{Multi-View Obs.} whether the dataset includes
multi-view robot observations. \textbf{Gaze Freq.}, \textbf{Gaze Acq.}, and
\textbf{Set.} denote eye-tracker frequency, acquisition setup, and data setting (Real or Simulated).
}
\vspace{-2mm}
\label{tab:dataset_comparison}
\centering
\footnotesize
\setlength{\tabcolsep}{3.8pt}
\renewcommand{\arraystretch}{1.10}

\begin{tabular*}{\textwidth}{@{\extracolsep{\fill}}lccccccccccc@{}}
\hline
\textbf{Dataset}
& \textbf{\shortstack{Recorded\\Human Gaze}}
& \textbf{\shortstack{Robot-View\\Gaze}}
& \textbf{\shortstack{Multi-View\\Gaze}}
& \textbf{\shortstack{Gaze\\Freq.}}
& \textbf{\shortstack{Gaze\\Acq.}}
& \textbf{\shortstack{Multi-View\\Obs.}}
& \textbf{Demos}
& \textbf{Tasks}
& \textbf{\shortstack{Avg.\\Dur. (s)}}
& \textbf{Participant}
& \textbf{Set.} \\
\hline

HARMONIC~\cite{newman2020harmonicmultimodaldatasetassistive}
& \cmark & \xmark & \xmark & $120$ Hz & HMD
& \xmark & $480$ & $1$ & $37.5$ & $24$ & Real \\

RoboTurk-SIM~\cite{mandlekar2018roboturkcrowdsourcingplatformrobotic}
& \xmark & \xmark & \xmark & -- & --
& \cmark & $3{,}224$ & $3$ & $153.5$ & n/a & Sim \\

RoboTurk-Real~\cite{mandlekar2019scalingrobotsupervisionhundreds}
& \xmark & \xmark & \xmark & -- & --
& \cmark & $2{,}144$ & $3$ & $187.2$ & $54$ & Real \\

BridgeData V2~\cite{walke2024bridgedatav2datasetrobot}
& \xmark & \xmark & \xmark & -- & --
& \cmark & $60{,}096$ & $13$ & $7.6$ & n/a & Real \\

DROID~\cite{khazatsky2025droidlargescaleinthewildrobot}
& \xmark & \xmark & \xmark & -- & --
& \cmark & $76{,}000$ & $86$ & $16.6$ & $50$ & Real \\

RT-1~\cite{brohan2023rt1roboticstransformerrealworld}
& \xmark & \xmark & \xmark & -- & --
& \cmark & $130{,}000$ & $744$ & n/a & n/a & Real \\

DAA~\cite{kim2024multitaskrealrobotdatagaze}
& \cmark & \cmark & \xmark & $120$ Hz & HMD
& \xmark & $224{,}210$ & $11$ & $2.4$ & n/a & Real \\

\hline
\textbf{\methodname{} (Ours)}
& \cmark & \cmark & \cmark & $300$ Hz & Screen
& \cmark & $12{,}000$ & $40$ & $30.0$ & $20$ & Real \\
\hline

\end{tabular*}
\vspace{-6pt}
\end{table*}

% =====================================================================
%  3.4 GAZE-GUIDED ATTENTION REGULARIZATION (method contribution)
% =====================================================================
\subsection{Attention Regularization}
\label{sec:reg}

The patch-level gaze distribution $G_t^{i}$ in Eq.~\eqref{eq:pool} represents the
regions human operators attended to during execution, expressed over the same
visual-token grid used by the policy. We use it to regularize the policy's
internal language-conditioned visual attention during training
(Figure~\ref{fig:arch}). Rather than forcing the model to imitate human gaze
exactly, the objective encourages attention toward regions relevant to the
demonstrated action.

Let $X_{\mathrm{vis}}^{i}\in\mathbb{R}^{N_v\times d}$ and
$X_{\mathrm{lang}}\in\mathbb{R}^{N_l\times d}$ denote the visual and language
tokens, where $N_v$ and $N_l$ are the numbers of visual and language tokens and
$d$ is the embedding dimension. We extract attention from the final vision-language fusion layer, where a
language-conditioned query $Q_{\mathrm{lang}}$ attends over the visual keys
$K_{\mathrm{vis}}^{i}$ of view $i$:
\begin{equation}
S_t^{i}
= \mathrm{Softmax}\!\left(
\frac{Q_{\mathrm{lang}} (K_{\mathrm{vis}}^{i})^{\top}}{\sqrt{d}}\right)
\in \mathbb{R}^{1\times N_v}.
\label{eq:attn}
\end{equation}
$S_t^i[j]$ quantifies the importance assigned to the $j^{\text{th}}$ visual patch
in view~$i$ conditioned on the instruction, i.e., the model's estimate of
task-relevant tokens. We align this attention with the human gaze distribution
using a KL-divergence loss, combined with the standard action objective:
\begin{equation}
\mathcal{L}_{\mathrm{attention}}
= \lambda\, D_{\mathrm{KL}}\!\left(G_t^{i} \,\|\, S_t^{i}\right),
\qquad
\mathcal{L} = \mathcal{L}_{\mathrm{action}} + \mathcal{L}_{\mathrm{attention}}.
\label{eq:loss}
\end{equation}
The coefficient $\lambda$ controls the strength of regularization. During
training, $G_t^i$ is used only to compute $\mathcal{L}_{\mathrm{attention}}$; at
deployment, gaze is not needed and the model follows the same
image-language-proprioception forward path as the baseline VLAs.

%% file: sections/experiments.tex
\section{Experiments}
\label{sec:experiments}
%\vspace{-1mm}

We evaluate \methodname{} as both a gaze-equipped real-robot dataset and a
framework for using teleoperation gaze to regularize visual attention in VLA
policies. We use $\pi_{0.5}$~\cite{intelligence2025pi05visionlanguageactionmodelopenworld}
as the primary backbone. We first compare \methodname{} with representative robot
and gaze-augmented datasets to highlight the gap in real human gaze synchronized
with multi-view robot observations.We then evaluate attention regularization with gaze-derived targets through
real-world and robustness trials, regularization-strength and gaze-source
ablations, cross-architecture and simulation transfer, attention visualizations,
and quantitative gaze-transfer analysis. Gaze is used only during training;
inference uses the same images, language, and proprioception as the baseline.

\subsection{Comparison with Robot Manipulation and Gaze-Augmented Datasets}
\label{sec:dataset_comparison}

Table~\ref{tab:dataset_comparison} summarizes the data gap addressed by
\methodname{}. Existing large-scale robot datasets do not include
human gaze whereas gaze-augmented datasets record human gaze, but
often through head-mounted or egocentric setups that can be intrusive, sensitive
to calibration drift, and are limited to a single displayed view. As a result, they
do not directly provide gaze supervision across the multi-view robot
observations consumed by modern VLA policies.

\methodname{} addresses this gap by recording gaze directly on the robot camera
feed used during teleoperation, logging the active view, and propagating gaze to
the remaining synchronized camera views through cross-view alignment. This yields
robot-view gaze on the operator-selected view and aligned gaze supervision
across the policy's multi-view visual input. The dataset further includes
hardware TTL synchronization, five-point screen calibration before each
participant/day shift, $300$Hz gaze recording, $40$ tasks across five domains,
$20$ participants, and approximately $30\,\mathrm{s}$ trajectories.

\subsection{Real-World Evaluation}
\label{sec:expanded-real-world}

We evaluate whether attention regularization with real teleoperation gaze
improves VLA policy learning in a physical robot setting. All policies are trained for $40$k steps under the matched
conditions, using the same task demonstrations and architecture; the only difference is whether gaze-derived targets are used for attention
regularization during training. At inference time, both policies receive the same inputs:
images, language, and proprioception.

For each task and method, we run $100$ real-world rollouts, split evenly between
$50$ standard trials and $50$ robustness trials. Standard trials use the common
task configuration from data collection. Robustness trials preserve the same
language instruction and success criterion, but introduce controlled spatial or
visual changes. For example, in the cup-placement task, the cup and plate may be
moved to new locations, nearby objects may change appearance, or unseen
distractors may be added. This protocol tests whether gaze supervision improves
not only execution under familiar conditions, but also visual grounding when the
policy must identify task-relevant objects under scene variation.

Table~\ref{tab:gaze_results} reports standard, robustness, and overall success
across all $100$ rollouts, with Wilson confidence intervals for the overall
success rate. Attention regularization with gaze-derived targets improves average standard success from
$55.8\%$ to $66.8\%$, robustness success from $40.3\%$ to $50.3\%$, and overall
success from $48.1\%$ to $58.6\%$, corresponding to a $10.5$ percentage-point
overall gain. The attention-regularized policy improves over the baseline on every
task. These results provide real-robot evidence that
teleoperation gaze can serve as an effective training-time supervision signal
for improving task-relevant visual grounding in VLA manipulation policies.

%\vspace{2mm}
\begin{table*}[t]
\caption{
Expanded real-world evaluation with $100$ rollouts per task and method.
Standard and robustness columns report successes out of $50$ trials; overall
columns report successes out of $100$ trials with $95\%$ Wilson confidence
intervals. Improvement is the absolute percentage-point gain in overall success.
}
%\vspace{-2mm}
\label{tab:gaze_results}
\centering
\footnotesize
\setlength{\tabcolsep}{3pt}
\renewcommand{\arraystretch}{1.08}

\begin{tabular*}{\textwidth}{@{\extracolsep{\fill}}lccccccc@{}}
\toprule
& \multicolumn{2}{c}{\textbf{Standard}}
& \multicolumn{2}{c}{\textbf{Robustness}}
& \multicolumn{2}{c}{\textbf{Overall}}
& \textbf{Gain} \\
\cmidrule(lr){2-3}
\cmidrule(lr){4-5}
\cmidrule(lr){6-7}
\textbf{Task}
& \textbf{Base}
& \textbf{Ours}
& \textbf{Base}
& \textbf{Ours}
& \textbf{Base [95\% CI]}
& \textbf{Ours [95\% CI]}
& \textbf{pp} \\
\midrule
Pick cup and place on plate
& 25/50 & \textbf{37/50} & 14/50 & \textbf{27/50}
& 39 [30.0,48.8] & \textbf{64 [54.2,72.7]} & \textbf{+25} \\

Pick cup and place inside container
& 36/50 & \textbf{40/50} & 30/50 & \textbf{33/50}
& 66 [56.3,74.5] & \textbf{73 [63.6,80.7]} & \textbf{+7} \\

Pick up towel and place in basket
& 42/50 & \textbf{47/50} & 36/50 & \textbf{43/50}
& 78 [68.9,85.0] & \textbf{90 [82.6,94.5]} & \textbf{+12} \\

Pick up cube and place on plate
& 40/50 & \textbf{44/50} & 33/50 & \textbf{38/50}
& 73 [63.6,80.7] & \textbf{82 [73.3,88.3]} & \textbf{+9} \\

Open rice cooker
& 34/50 & \textbf{41/50} & 21/50 & \textbf{30/50}
& 55 [45.3,64.4] & \textbf{71 [61.5,79.0]} & \textbf{+16} \\

Pour contents into rice cooker
& 41/50 & \textbf{44/50} & 37/50 & \textbf{40/50}
& 78 [68.9,85.0] & \textbf{84 [75.6,89.9]} & \textbf{+6} \\

Transfer tennis ball between containers
& 9/50 & \textbf{19/50} & 6/50 & \textbf{9/50}
& 15 [9.3,23.3] & \textbf{28 [20.1,37.5]} & \textbf{+13} \\

Hit nail on stand with hammer
& 37/50 & 37/50 & 25/50 & \textbf{28/50}
& 62 [52.2,70.9] & \textbf{65 [55.3,73.6]} & \textbf{+3} \\

Turn stove-top knob
& 26/50 & \textbf{30/50} & 21/50 & \textbf{25/50}
& 47 [37.5,56.7] & \textbf{55 [45.3,64.4]} & \textbf{+8} \\

Pick up injection and place on tray
& 28/50 & \textbf{36/50} & 18/50 & \textbf{25/50}
& 46 [36.6,55.7] & \textbf{61 [51.2,70.0]} & \textbf{+15} \\

Place colored cup in matching stand
& 32/50 & \textbf{37/50} & 21/50 & \textbf{24/50}
& 53 [43.3,62.5] & \textbf{61 [51.2,70.0]} & \textbf{+8} \\

Place red pill in pill case
& 24/50 & \textbf{30/50} & 17/50 & \textbf{24/50}
& 41 [31.9,50.8] & \textbf{54 [44.3,63.4]} & \textbf{+13} \\

Pick up tennis balls and place into basket
& 14/50 & \textbf{19/50} & 8/50 & \textbf{12/50}
& 22 [15.0,31.1] & \textbf{31 [22.8,40.6]} & \textbf{+9} \\

Place inverted test tube on stand
& 26/50 & \textbf{30/50} & 16/50 & 16/50
& 42 [32.8,51.8] & \textbf{46 [36.6,55.7]} & \textbf{+4} \\

Arrange knife and fork in cutlery case
& 14/50 & \textbf{20/50} & 8/50 & \textbf{15/50}
& 22 [15.0,31.1] & \textbf{35 [26.4,44.7]} & \textbf{+13} \\

Hammer correct large knob
& 34/50 & \textbf{37/50} & 20/50 & \textbf{25/50}
& 54 [44.3,63.4] & \textbf{62 [52.2,70.9]} & \textbf{+8} \\

Transfer book from stand to shelf
& 27/50 & \textbf{35/50} & 22/50 & \textbf{28/50}
& 49 [39.4,58.7] & \textbf{63 [53.2,71.8]} & \textbf{+14} \\

Turn on sink faucet
& 13/50 & \textbf{18/50} & 10/50 & \textbf{11/50}
& 23 [15.8,32.2] & \textbf{29 [21.0,38.5]} & \textbf{+6} \\

\midrule
\textbf{Average}
& 27.9/50 & \textbf{33.4/50}
& 20.2/50 & \textbf{25.2/50}
& 48.1 [45.8,50.4]
& \textbf{58.6 [56.3,60.8]}
& \textbf{+10.5} \\
\bottomrule
\end{tabular*}
%\vspace{-6pt}
\end{table*}

\subsection{Effect of Attention Regularization}
\label{sec:gaze-effect}

Gaze-guided attention regularization is the mechanism by which \methodname{}
turns real robot-view human gaze into training-time supervision. To understand when and why gaze-derived targets improve policy learning through
attention regularization, we isolate two factors:
the strength of the regularization loss and the source of the gaze target. This
allows us to distinguish improvements from real human gaze as supervision targets from improvements
that may come from generic attention shaping.

\textbf{Regularization Strength.}
\label{sec:gaze-regularization-strength}
The coefficient $\lambda$ controls how strongly gaze supervision influences
policy attention during training (Table~\ref{tab:gaze_ablation_combined}, top).
A low value $(\lambda{=}0.001)$ performs best across all tasks and improves
over the baseline in every case, while strong regularization
$(\lambda{=}10)$ falls below the baseline on all tasks. This suggests that gaze
is most useful as a soft supervision signal that guides visual grounding without
overpowering the action-learning objective.

\textbf{Regularization Source.}
\label{sec:gaze-regularization-source}
We next test whether the improvement comes from real teleoperation gaze rather
than from generic attention shaping. We compare attention-regularized policies
trained with four target sources: pseudo-gaze targets from an off-the-shelf
saliency estimator~\cite{lai2023eye}, uniform targets that assign equal mass to
all visual patches, central targets that concentrate mass near the image center,
and real human gaze recorded during teleoperation. Table~\ref{tab:gaze_ablation_combined}
(bottom) shows that pseudo-gaze targets improve some tasks, but real human gaze
provides the strongest attention targets and achieves the highest success on
every task. This suggests that human fixations provide task-relevant attention
supervision beyond generic, central, or saliency-derived priors.

\begin{table}[t]
\centering
\footnotesize
\setlength{\tabcolsep}{3pt}
\renewcommand{\arraystretch}{1}

\caption{
\textbf{Top:} varying the regularization strength $\lambda$ shows
that gaze-derived target is most effective as a soft prior. \textbf{Bottom:} comparing
supervision sources shows that real human gaze provides the strongest attention targets.
}
%\vspace{-2mm}
\label{tab:gaze_ablation_combined}

% ---------------- Top block ----------------
\begin{tabular*}{0.94\linewidth}{@{\extracolsep{\fill}}lcccc@{}}
\toprule
\textit{Task Instruction}
& \textbf{Base}
& \multicolumn{3}{c}{\textbf{Ours}} \\
\cmidrule(lr){2-2} \cmidrule(lr){3-5}
& \textcolor{gray}{$\lambda{=}0$}
& \textcolor{gray}{0.001}
& \textcolor{gray}{0.01}
& \textcolor{gray}{10} \\
\midrule
Pick cup and place on plate         & 39 & \textbf{64} & 42 & 24 \\
Pick cup and place inside container & 66 & \textbf{73} & 63 & 30 \\
Pick up towel and place in basket   & 78 & \textbf{90} & 72 & 40 \\
Pick up cube and place on plate     & 73 & \textbf{82} & 77 & 56 \\
Pick up injection place on tray     & 46 & \textbf{61} & 41 & 33 \\
Open rice cooker                    & 55 & \textbf{71} & 53 & 45 \\
\midrule
\textbf{Average}                    & 59.5 & \textbf{73.5} & 58.0 & 38.0 \\
\bottomrule
\end{tabular*}

%\vspace{6pt}

% ---------------- Bottom block ----------------
\begin{tabular*}{0.94\linewidth}{@{\extracolsep{\fill}}lcccc@{}}
\toprule
\textit{Task Instruction}
& \textbf{Pseudo}
& \textbf{Uniform}
& \textbf{Central}
& \textbf{Real} \\
\midrule
Pick cup and place on plate         & 58 & 36 & 43 & \textbf{64} \\
Pick cup and place inside container & 69 & 57 & 59 & \textbf{73} \\
Pick up towel and place in basket   & 80 & 62 & 73 & \textbf{90} \\
Pick up cube and place on plate     & 77 & 57 & 63 & \textbf{82} \\
Pick up injection place on tray     & 56 & 38 & 44 & \textbf{61} \\
Open rice cooker                    & 63 & 47 & 51 & \textbf{71} \\
\midrule
\textbf{Average}                    & 67.2 & 49.5 & 55.5 & \textbf{73.5} \\
\bottomrule
\end{tabular*}
\vspace{-8mm}
\end{table}

\subsection{Attention Visualization}
\begin{figure}[t]
  \centering
  \includegraphics[width=0.98\linewidth]{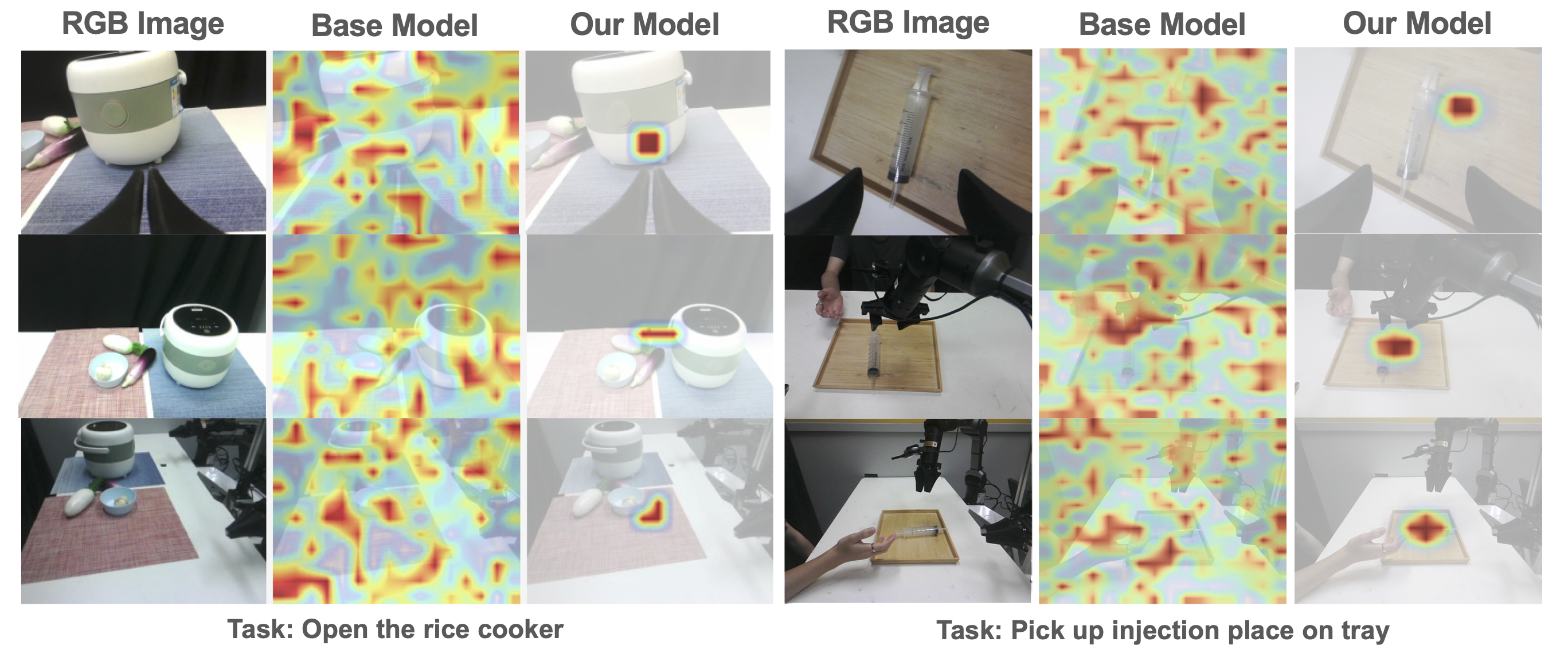}
  %\vspace{-2mm}
  \caption{
\textbf{Attention Visualization.} Qualitative attention maps for rice-cooker and injection-tray rollouts. Warmer
regions indicate higher language-conditioned visual attention. Compared with the
baseline, the attention-regularized policy places more attention on actionable
regions, including the rice-cooker button and the injection tool, while using no
gaze input at inference. }
  \label{fig:attention}
    \vspace{-8pt}
\end{figure}

To better understand how attention regularization affects the learned policy, we
visualize language-conditioned attention over visual patches
(Eq.~\eqref{eq:attn}) for the baseline and the attention-regularized policy on
matched scenes (Fig.~\ref{fig:attention}). These visualizations help interpret
how gaze-derived attention targets improve task-relevant visual grounding. The
baseline attention is often diffuse or partially attracted to background
regions, while the attention-regularized policy produces more concentrated
attention on objects, contact points, and target regions that are directly
relevant for action execution. In the ``open rice cooker'' example, the
attention-regularized policy attends more strongly to the button and interaction
area; in the injection-tray example, it focuses on the injection syringe rather
than spreading attention diffusely across the scene.

% --------------------------- FIGURE (Attention-visualisation) --------------------
%\vspace{2mm}

% ---------------------------------------------------------------------

\subsection{Generalization Across Settings and Architectures}
\label{sec:generalization}
\textbf{Performance on other VLA architectures.}
To test whether the effect of using gaze as supervision for attention regularization transfers across VLA
architectures, we apply the same attention regularization objective to OpenVLA~\cite{kim2024openvlaopensourcevisionlanguageactionmodel}, while keeping
the task set and training data fixed. The goal is
not to rank the two architectures, but to test whether the same training signal
provides consistent gains beyond a single backbone.
Table~\ref{tab:cross_arch} reports results on six real-world tasks that cover
precise interaction, object transfer, and deformable-object placement. Attention
regularization improves success rates for both $\pi_{0.5}$ and OpenVLA on each task.
For $\pi_{0.5}$, gains range from 9 to 25 percentage points; for OpenVLA, gains range from 7 to 14 percentage points. 
These results suggest that attention regularization with gaze-derived targets
functions as an architecture-agnostic training signal, improving visual grounding across different VLA backbones
without changing the inference interface.

\begin{table}[t]
  \centering
  \caption{Cross-architecture evaluation on real-world tasks.
    Success rates (\%) for $\pi_{0.5}$ and OpenVLA with and without attention
    regularization. Best result per task and backbone is in \textbf{bold}.}
    %\vspace{-2mm}
  \label{tab:cross_arch}
  \small
  \setlength{\tabcolsep}{4pt}
  \renewcommand{\arraystretch}{1.1}
  \begin{tabular}{@{}l cc cc@{}}
    \toprule
    & \multicolumn{2}{c}{$\pi_{0.5}$}
    & \multicolumn{2}{c}{OpenVLA} \\
    \cmidrule(lr){2-3}\cmidrule(lr){4-5}
    \textit{Task}                 & \textbf{Base} & \textbf{Ours}
                                  & \textbf{Base} & \textbf{Ours} \\
    \midrule
    Open rice cooker              & 55 & \textbf{71} & 47 & \textbf{60} \\
    Transfer tennis balls         & 15 & \textbf{28} & 10 & \textbf{17} \\
    Place towel in basket         & 78 & \textbf{90} & 63 & \textbf{77} \\
    Pick cube, place on plate     & 73 & \textbf{82} & 57 & \textbf{70} \\
    Pick cup, place on plate      & 39 & \textbf{64} & 51 & \textbf{60} \\
    Pick injection, place on tray & 46 & \textbf{61} & 43 & \textbf{50} \\
    \midrule
    \textbf{Average}              & 51.0 & \textbf{66.0} & 45.2 & \textbf{55.7} \\
    \bottomrule
  \end{tabular}
  \vspace{-5mm}
\end{table}

\textbf{Simulation Performance.}
We further test whether real gaze supervision transfers beyond the physical
ALOHA setup using LIBERO-Spatial~\cite{liu2023libero}, a simulated manipulation
benchmark with different scene layouts and objects. We collect gaze with the
same screen-based setup used in the real-robot experiments: participants observe
task demonstrations on a display while the eye tracker records gaze on the
shown camera view. In this simulated setting, demonstrations are observed from a
single view, and the resulting gaze heatmaps are used as training-time
supervision under the same protocol as the real-world experiments. We compare
against a baseline trained without gaze.
As shown in Table~\ref{tab:libero_spatial_real_gaze}, attention regularization
improves overall success from $57.9\%$ to $69.6\%$, with
improvements across all ten spatial configurations. This suggests that gaze-derived attention targets remain useful beyond the physical robot setup, even when the
environment, camera setting, and visual coverage differ from the real-world
collection setup.

%\vspace{2mm}
\begin{table}[t]
\centering
\footnotesize
\setlength{\tabcolsep}{3pt}
\renewcommand{\arraystretch}{1}
\caption{
LIBERO-Spatial success rates (\%) at $30$k steps. Real human gaze as supervision for attention regularization improves
performance across all spatial configurations.
}
%\vspace{-2mm}
\label{tab:libero_spatial_real_gaze}
\begin{tabular}{@{}>{\raggedright\arraybackslash}p{0.60\linewidth}cc@{}}
\toprule
\textbf{Object Location} & \textbf{Base} & \textbf{Ours} \\
\midrule
Between plate and ramekin & 83.0 & \textbf{90.3} \\
Next to ramekin           & 73.0 & \textbf{84.0} \\
Table center              & 83.0 & \textbf{90.7} \\
On cookie box             & 64.0 & \textbf{74.0} \\
In cabinet drawer         & 62.0 & \textbf{72.0} \\
On ramekin                & 36.0 & \textbf{50.0} \\
Next to cookie box        & 62.0 & \textbf{72.0} \\
On stove                  & 19.0 & \textbf{35.0} \\
Next to plate             & 34.0 & \textbf{55.0} \\
On wooden cabinet         & 63.0 & \textbf{73.0} \\
\midrule
\textbf{Overall Avg.}     & 57.9 & \textbf{69.6} \\
\bottomrule
\end{tabular}
\vspace{-4mm}
\end{table}

%\vspace{-4mm}
\subsection{Quantitative Analysis of Gaze Transfer}
\label{sec:gaze_transfer_quant}

We performed a manual ground-truth evaluation of cross-view gaze transfer on
$300$ synchronized frames. An annotator viewed the source-camera image with
the recorded gaze location and marked the corresponding location in the
synchronized target view. Since gaze supervision is applied at the patch
level, we evaluate the distance to the manually annotated ground truth in
visual patches. Geometry alone produced a mean error of $2.57$ patches, while
our geometry-guided hybrid reduced this to just $1.36$ patches, showing that
the transferred gaze remains close to the ground-truth patch used for
supervision.

We also measure transfer reliability across five representative tasks. On
average, the hybrid geometry and LightGlue transfer succeeds for $71.0\%$ of
right-wrist targets and $79.4\%$ of left-wrist targets. When reliable
correspondences cannot be found because of occlusion or limited view overlap, the
method falls back to a center prior. Because \methodname{} retains the original
source-view gaze and synchronized camera streams, gaze transfer can be
recomputed with future registration methods without recollecting demonstrations.

%% file: sections/conclusion.tex
\section{Conclusion}
\label{sec:conclusion}

% We introduced \methodname{}, a gaze-equipped manipulation dataset and training
% framework for improving visual grounding in transformer-based VLA policies.
% \methodname{} records human gaze on the robot camera view used during
% teleoperation, synchronizes it with multi-view robot observations and actions,
% and propagates it across views through cross-view alignment. This provides
% gaze-derived targets for the visual inputs used by multi-view policies,
% addressing a key limitation of prior gaze-equipped robot datasets in which gaze
% is typically tied to a single view.
% Using these targets, \methodname{} regularizes language-conditioned visual
% attention during training while preserving the standard gaze-free deployment
% interface. Across real-world ALOHA tasks, ablations, simulation transfer,
% attention visualizations, and cross-architecture evaluation, attention
% regularization with real human gaze improves visual grounding and manipulation
% success. These results show that human gaze, when converted into attention
% targets, can serve as a practical training-time supervision signal for multi-view
% VLA policy learning.

We introduced \methodname{}, a gaze-equipped manipulation dataset and training
framework for improving visual grounding in transformer-based VLA policies.
\methodname{} records human gaze on the robot camera view used during
teleoperation, aligns it with multi-view robot observations and actions, and
propagates it across views through cross-view alignment. This provides
gaze-derived targets for multi-view policy inputs, addressing a key limitation of
prior gaze-equipped robot datasets in which gaze is typically tied to a single
view. Using these targets, \methodname{} regularizes language-conditioned visual
attention during training while preserving gaze-free deployment. Across
real-world ALOHA tasks, ablations, simulation transfer, attention visualizations,
and cross-architecture evaluation, attention regularization with real human gaze
improves visual grounding and manipulation success. These results show that
human gaze, when converted into attention targets, can serve as a practical
training-time supervision signal for multi-view VLA policy learning.